\documentclass{bmvc2k}
\usepackage{multirow}

\usepackage{xcolor}

\title{NSSR-DIL: Null-Shot Image Super-Resolution Using Deep Identity Learning}

\addauthor{Sree Rama Vamsidhar S}{sreeramvds@gmail.com}{1}
\addauthor{Rama Krishna Gorthi}{rkg@iittp.ac.in}{1}

\addinstitution{
 Indian Institute of Technology (IIT) Tirupati\\
 India
}

\runninghead{SR Vamsidhar, RK Gorthi}{NSSR-DIL}

\begin{document}
\maketitle
\begin{abstract}
The present State-of-the-Art (SotA) Image Super-Resolution (ISR) methods employ Deep Learning (DL) techniques using a large amount of image data. The primary limitation to extending the existing SotA ISR works for real-world instances is their computational and time complexities. In this paper, contrary to the existing methods, we present a novel and computationally efficient ISR algorithm that is independent of the image dataset to learn the ISR task. The proposed algorithm reformulates the ISR task from generating the Super-Resolved (SR) images to computing the inverse of the kernels that span the degradation space. We introduce ``Deep Identity Learning”, exploiting the identity relation between the degradation and inverse degradation models. The proposed approach neither relies on the ISR dataset nor on a single input low-resolution (LR) image (like the self-supervised method i.e. ZSSR) to model the ISR task. Hence we term our model as ``Null-Shot Super-Resolution Using Deep Identity Learning (NSSR-DIL)". The proposed NSSR-DIL model requires fewer computational resources, at least by an order of 10, and demonstrates a competitive performance on benchmark ISR datasets. Another salient aspect of our proposition is that the NSSR-DIL framework detours retraining the model and remains the same for varying scale factors like $\times 2, \times3, \times4$. This makes our highly efficient ISR model more suitable for real-world applications.
\end{abstract}
\section{Introduction}
\label{sec:intro}
Super Resolution is a well-established low-level vision task whose objective is to generate a High-Resolution (HR) image from the given corresponding LR observation(s). Real-world applications in prominent domains like medical imaging, satellite imaging, and surveillance demand the HR version of the scene of interest for its analysis and understanding.
\par In 2014, the first DL-based ISR work i.e., SRCNN \cite{srcnn} had demonstrated remarkable improvement over the existing dominant example-based ISR methods \cite{example1, sisr, example3}. Further, various deep learning techniques \cite{lightsr, LapSRN, lightisr-spl,srgan, esrgan, edsr, attn1, attn2} were proposed to generate the super-resolved images with finer details and of better quality. These works consider the unknown degradation process as a learnable entity in either supervised or unsupervised approach from a large synthetic HR-LR image pair or LR image datasets respectively. The existing works \cite{srcnn, edsr, attn1, attn2, LapSRN} are non-blind as they presume the degradation model in their dataset to be MATLAB bicubic downscaling or some blurring kernel with additive or multiplicative noise followed by downscaling operation. Further, a line of works like \cite{srmd, udvd, fssr, cincgan, bsrgan} considered multiple types of degradations in their datasets and illustrated the improved performance over a wide range of degradations in LR images. \\ 
Another line of works address the fundamental problem in ISR algorithms i.e., the domain gap between the synthetic and the Real LR (RLR) images by acquiring the realistic datasets. The real HR-LR image pairs datasets were captured with various techniques like the focal length adjusting-based approach \cite{cai2019toward, wei2020component, chen2019camera, zhang2019zoom}, the hardware binning-based approach \cite{kohler2019toward}, and the beam splitter-based approach \cite{joze2020imagepairs}. However, misalignment between LR and HR image pairs despite image registration introduces blurring artifacts in the reconstructed HR images. Moreover, capturing and preparing a realistic dataset is a hindering task requiring lots of human effort, time, and, computational resources. Further, it is not scalable for the generality of all types of real-world degradations.\\
The other stream of works are zero-shot approaches \cite{zssr, kernelgan, dbpi, dualsr}, that attempt LR image-specific super-resolution. These works are training dataset-independent and rely on internal statistics of an image i.e., patch recurrence property \cite{sisr}. The primary shortcoming of the existing zero-shot approaches is that the inference time taken for a single image is very high and is not suitable for practical applications. 
\par In this paper, we redefine the task of learning the ISR model from image data to computing an inverse of the degradation model. Here, the degradation model is constructed using a wide set of anisotropic Gaussian kernels ($K$s). We propose a novel learning strategy called ``Deep Identity Learning (DIL)" that exploits the identity relation between the degradation model and its inverse ($K^{-1}$) i.e., the ISR model. In this work, a custom lightweight and computationally efficient CNN with no activation functions, referred to as Linear-CNN (L-CNN), is trained on the generated degradation model with DIL as its objective. Unlike existing self-supervised zero-shot ISR frameworks \cite{zssr}, \cite{dbpi}, and \cite{dualsr} the proposed NSSR-DIL model is image independent and doesn't require LR-HR pair or LR image data at any stage of the ISR task learning. The highlights of our proposition in this paper are as follows. 
\begin{enumerate}
    \item To the best of our knowledge, NSSR-DIL is the first image data-independent ISR model and is contrary to the mainstream SoTA works that do “distribution mapping from LR to HR image data” using DL frameworks.
    \item  A novel learning objective exploiting the identity relation between the degradation model and its inverse i.e., DIL, is proposed.
    \item The proposed ISR model is at least ten times more computationally efficient and delivers competitive ISR performance.     
\end{enumerate} 
\section{Related work}
In this section, the relevant DL-based zero-shot ISR methods were discussed in brief.\\
A stream of works explores the internal statistics based on the recurrence property of a natural image to model the degradation from a given single test input, the LR image itself. The recurrence property of a natural image states that patches of a single image tend to recur within and across different scales of this image. Glasner et al. \cite{sisr} proposed to capitalize the internal statistics within an image to tackle the Single ISR problem. Non-parametric blind super-resolution \cite{npbsr} utilized this framework for the blind ISR task. Based on this framework, Zero-Shot Super-Resolution (ZSSR) \cite{zssr} proposed to train an image-specific CNN with HR-LR pairs generated from a single LR input, for super-resolving the same input LR image. KernelGAN \cite{kernelgan} was proposed for blind kernel estimation using the patch recurrence property. For a given arbitrary LR image, the kernel recovery performance is limited and unstable. In recent times, Reference-based Zero-Shot Super-Resolution with Depth Guided Self-Exemplars (RZSR) \cite{rzsr} was proposed based on ZSSR \cite{zssr} and reference-based ISR techniques and also utilizes the predicted depth information of the given LR image in its ISR framework. Flow-based kernel prior (FKP) \cite{liang21fkp}, was proposed for robust kernel estimation, based on Normalizing flow (NF) \cite{dinh2014nice}, \cite{dinh2016density} to learn a kernel prior in latent space. Nonetheless, these kernel estimation works are to be associated with the existing ISR models to generate the super-resolved image. Later, Kim et al. \cite{dbpi} proposed a unified internal learning-based SR framework consisting of an SR network and a downscaling network. In the self-supervised training phase of DBPI, the SR network is optimized to reconstruct the LR input image from its downscaled version produced by the downscaling network. Meanwhile, the downscaling network is trained to recover the LR input image from its super-resolved version generated by the SR network. Similarly, Emad et al. \cite{dualsr} proposed DualSR: Zero-Shot Dual Learning for Real-World Super-Resolution, which jointly optimizes an image-specific down sampler and corresponding upsampler with the cycle-consistency loss, the masked interpolation loss, and the adversarial loss using the patches from the test image. The most recent zero-shot ISR method, Zero-Shot Dual-Lens Super-Resolution \cite{zeroduallens} learns an image-specific SR model from the dual-lens pair given at test time. \\ In this work, towards addressing the limitations of data dependency, time, and computational complexities for the practical use case of real-time implementation, we proposed a lightweight yet robust purely image data-independent model to restore the distorted LR image.
\begin{figure*}[t]
\centering
\includegraphics[width=0.9\textwidth, height = 4.75cm ]{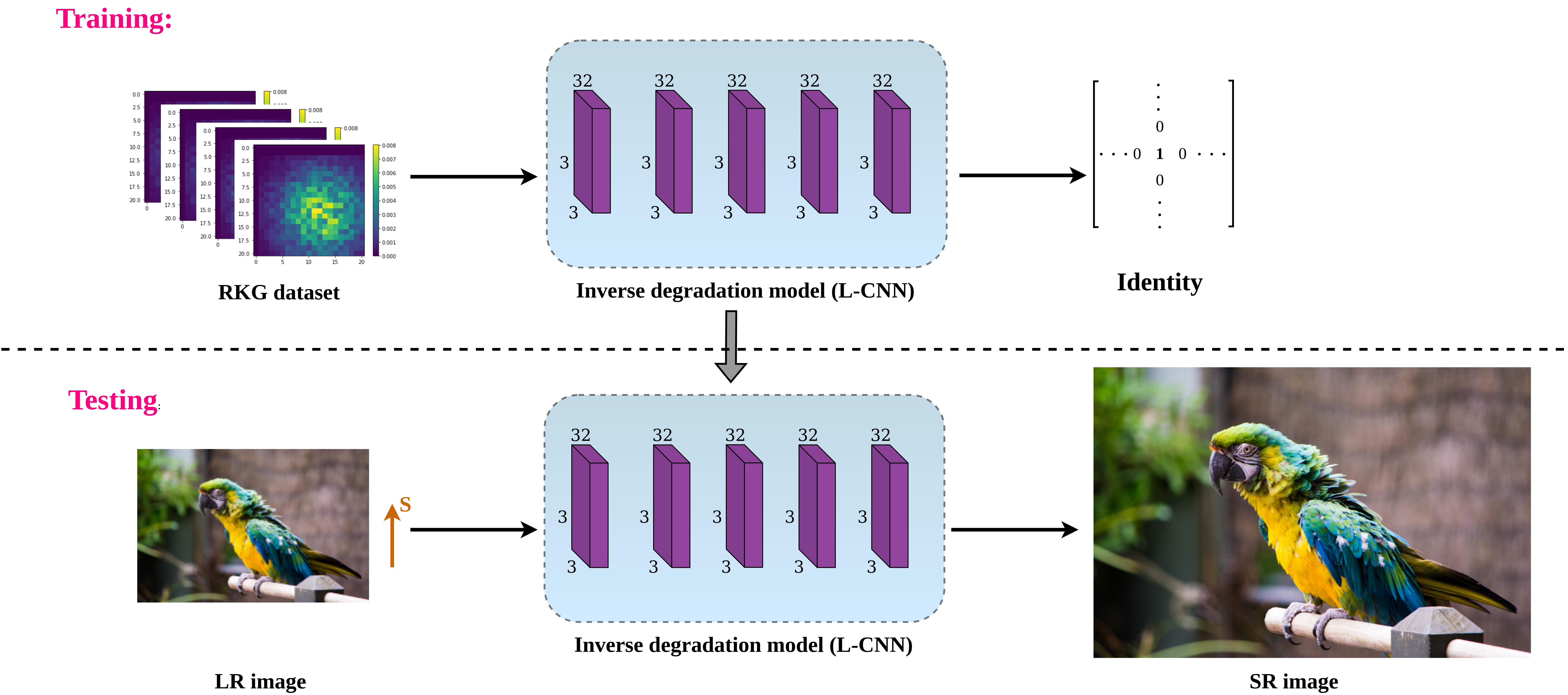}
\caption{{The training and inference methodology of the proposed NSSR-DIL model.}}
\label{fig:NSSR-DIL}
\end{figure*}
\section{Proposed method}
\label{proposed}
In this section, we elaborate on the proposed NSSR-DIL model in greater detail.\\
{\color{blue}{\textit{Problem introduction:}}} The end-to-end ISR framework can be broadly observed as the degradation model followed by the inverse degradation model. At first, the degradation model generates a degraded and downsampled version of the HR image i.e., the LR image. Later, the inverse degradation model restores the HR image from the obtained LR image i.e., the SR image. The degradation and inverse degradation models are discussed below in detail.\\
{\color{blue}{\textit{Degradation model:} }}

The degradation model typically consists of two stages. The first is the convolution of the HR image with the degradation kernel, followed by the down-sampling operation in the second stage. The mathematical representation of the correspondence between HR image $X \in {R}^{M \times N}$ and LR image $y \in {R}^{m \times n}$ is given in Eq. (\ref{eq1}) \cite{kernelgan}. 
\begin{equation}
    \label{eq1}
    y = (X \ast K) \downarrow_s 
\end{equation}
Here $K$ represents the degradation kernel, $\ast$ represents the convolution operation, $s$ is the scale factor (sf) and the dimension $M \times N ~\mbox{is equal to}~ sm \times sn$.
\\{\color{blue}{\textit{Inverse degradation model:}}}
The objective of the inverse of the degradation model is to reconstruct the HR image from the given LR image input. The mathematical representation of the equivalent inverse model of the degradation model is given in Eq. (\ref{eq2}).  
\begin{equation}
    \label{eq2}
 (y\uparrow^s) \ast K^{-1} = X
\end{equation}
Where $K^{-1}$ represents the inverse degradation kernel, to be estimated.
\\ {\color{blue}{\textit{Unified ISR model}}}. The unified degradation and inverse degradation models framework represents an ``identity model'' with the same entity as the input and also as the output i.e. HR image (in the ideal case). Here the degradation and inverse degradation models are characterized and effectively represented by their respective kernels $K$ and $K^{-1}$. In practice, the output of the inverse model is the super-resolved (SR) image from the LR image. 
In the unified ISR framework, it is assumed that the degraded HR images before the downsampling stage are adequately band-limited such that the downsampling and the immediate upsampling stages can be ignored. Then, a simplified ISR framework with a degradation model directly followed by the inverse degradation model is obtained with an HR image as both input and output. Therefore, the convolution operation between $K$ and $K^{-1}$ should result in an identity relation. Henceforth, we propose to learn the inverse degradation model forming an identity relation with the degradation kernel. The identity relation is given in Eq. (\ref{eq3}).
\begin{equation}
    \label{eq3}
 K \ast K^{-1} =  ~\delta
\end{equation}
Where $\delta$ is a two-dimensional discrete impulse function.\\ 
Based on the derived simplified form of the unified ISR model, we propose a simple, computationally efficient, and novel framework that is independent of images to learn the ISR task. The proposed formulation simplifies the task of learning ISR problem from image datasets using deep architectures with several millions of learnable parameters, to the task of ``identity'' learning between the degradation kernel $K$ and the inverse degradation kernel $K^{-1}$ through a custom linear CNN (discussed in Sec.\ref{L-CNN}). 
\subsection{Random Kernel Gallery (RKG) Dataset}
\label{kdataset}
To realize the identity learning task, a set of degradation maps sampled from the degradation kernel space is generated. In ISR literature, the degradation modeling is mainly carried out conventionally by using Gaussian prior \cite{gkprior1, npbsr} and, in recent works, using complex deep models like KMSR \cite{kmsr}, KernelGAN \cite{kernelgan} with their own underlying assumptions in the degradation model ($K$) estimation. Especially for real LR images, theoretically, $K$ is modeled as unimodal \cite{npbsr} \&  Gaussian \cite{CRMnBISR}. The existing well-known blind ISR works like \cite{kernelgan, dbpi} consider the $K$ to be isotropic or anisotropic Gaussian models. Also, the work \cite{deconvAlgo} demonstrated that with an appropriate estimation algorithm, blind deconvolution can be performed even with a weak Gaussian prior. Following these widely adopted premises, in our ISR work, we constructed the degradation model with the anisotropic Gaussian kernel.\\
In this work, a set of anisotropic Gaussian kernels with dimensions $21 \times 21$, having $\sigma_1, \sigma_2 \in U[0.175, 6]$ and rotation angle $\theta \in U[0, \pi]$ were randomly generated as representative samples of the degradation kernel space. The number of $K$ samples required to effectively learn the inverse degradation kernel $K^{-1}$ is determined empirically as 3,200. The reference output to learn the inverse degradation model is the two-dimensional impulse function mentioned in Eq.\ref{eq3} and its dimensions are based on the size of $K$ \& $K^{-1}$. We term this dataset as the ``Random Kernel Gallery (RKG)'' dataset.\\ The proposed NSSR-DIL method is not limited to the ``RKG-dataset'' with anisotropic Gaussian kernels. NSSR-DIL can be easily re-trained on any other kernel distribution if proven to be a more accurate representation of the degradations kernel space in RLR images.
\subsection{Linear Convolutional Neural Network (L-CNN)}
\label{L-CNN}
ISR is an ill-posed inverse problem for which a unique inverse will not exist. Computing the direct (or pseudo) inverse of the provided input degradation kernel $K$ or a single-layer network to learn the inverse of the degradation kernel cannot serve the ISR task's objective. This is because a matrix/single layer accepts only one set of parameters/global minima with convex loss \cite{kernelgan}. Also, the $K$ can usually be a low-rank matrix. Further, it was empirically found that single-layer architecture does not converge to the correct solution \cite{choromanska2015loss}. Whereas the multi-layered linear networks have many good and equally valued local minima. This allows many valid optimal solutions to the optimization objective in the form of different factorizations of the same matrix \cite{kawaguchi2016deep}, \cite{saxe2014a}, \cite{arora2018optimization}. Following these research results, we propose a multi-layer Linear CNN (L-CNN) with no activations to learn the inverse degradation kernel, with the degradation kernels ($K$) from the RKG-dataset as its input. The proposed L-CNN is a computationally efficient architecture having a depth of five layers and a width of 32 with $3 \times 3$ filters across the depth. Here the L-CNN is chosen to be a pre-upsample network \cite{srcnn}, which maintains the same output dimension at every layer. Therefore, at the inference stage, L-CNN operates on traditional upsampled LR images of any desired scale factor to generate SR images with fine details. This implies that the inverse degradation kernel $K^{-1}$ remains the same for varying scale factors like $\times 2, \times 3, \times 4$.
\\ The general limitations of the pre-upsample networks like complexity, computational time, and memory, because of operations in high dimensional space, are not valid in this work as the input to the CNN is kernel $K$, a matrix of very small dimension (i.e., $21 \times 21$), compared to very high dimensional or 3-D tensor input images, in practice. The learning and inference methodology of the NSSR-DIL model is depicted in Fig. \ref{fig:NSSR-DIL}.
\subsection{Deep Identity Learning (DIL):} The proposed DIL objective is to train the L-CNN on the RKG dataset as given in Eq. (\ref{eq6}).

\begin{equation}
    \label{eq6}
  Loss(L) = ~ \lVert K \ast K^{-1} - ~\delta \rVert^2_2 + R 
\end{equation}

Here, $R$ is the proposed regularization term, defined in Eq. (\ref{eq8}), and is provided to constrain the solution space to have ISR capability, in addition to learn a valid inverse.

\begin{equation}
\label{eq8}
  R = \lambda_1 \times L_{ConvArea} +\lambda_2 \times L_{center}
\end{equation}
where,

\begin{enumerate}

    \item  $L_{ConvArea} = ~\mid 1 - \sum_{i,j}K^{-1}_{i,j}\mid$;  This term holds one of the important properties of the convolution operation i.e. the area property since the proposed ISR approach learns the inverse degradation model only from the convolution operation. The area property states that the product of the area under the two input signals is equal to the area under the convolution output signal. Here, the two input signals are the degradation kernel and its inverse i.e. $K$ and $K^{-1}$, while the output signal is a 2-D impulse function. The area property of the convolution operation is discussed in detail below.
       \\   
        The Fourier transform of a 2-D discrete Time-domain signal can be written as
       \begin{equation}
       \label{eq9}
       \mathcal{F}(K(m,n) = \hat{K}(p,q) = \sum_{m}^{}\sum_{n}^{} K(m,n) \exp^{-j{\frac{2\pi pm}{N_1}}+{\frac{2\pi qn}{N_2}}}
       \end{equation}
        Here, $\mathcal{F}(.)$ represent the Fourier transform operation, (m,n) and (p,q) represents the index locations of the degradation and inverse degradation kernel matrices in the Time-domain and Frequency-domain signals respectively, and $N_1, N_2$ represent the dimensions of degradation, inverse degradation kernel.
       \\
      Consider the identity relation given in Eq. \ref{eq3}, and apply the Fourier transform on both sides i.e., $\mathcal{F}(K * K^{-1}) = \mathcal{F}(\delta)$;
       \\  The above expression can be written as $\hat{K}(p,q). \hat{K^{-1}}(p,q) = 1$, since the Convolution operation in the Time domain is equivalent to multiplication in the Frequency domain (and vice-versa). We know that the Fourier transform of the 2-D impulse function is 1 i.e. $\mathcal{F}(\delta) = 1$. Here '.' represents the element-wise multiplication operation.     
       \\
        Keeping the values of $(p,q) = (0,0)$ in the Eq. \ref{eq9}, we arrive at another important property of the Convolution operation that states the area under the Time-domain signal is equal to the value at the origin in the Frequency-domain form of that signal. Therefore, $ \hat{K}(0, 0) = \sum_{m}^{}\sum_{n}^{} K(m,n)$.
        Here, the $\sum_{m}^{}\sum_{n}^{} K(m,n)$ is set to be 1 in our RKG dataset, since a degradation kernel. Therefore, to satisfy the relation  $\hat{K}(p,q).\hat{K^{-1}}(p,q) = 1$ when $(p,q)=(0,0)$, the area under the $K^{-1}$ is imposed to be 1 i.e. $\sum_{m}^{}\sum_{n}^{} K^{-1}(m,n) = 1$ and introduced as a regularizer in DIL objective.
    \item $L_{center} = ~\mid 1 - K.K{^{-1}}^T \mid$; This term encourages the output of the LCNN to have a unit value at the center. Considering the Eq.\ref{eq3}, $[K * K^{-1}](m,n) = \delta(m,n) = 1, if (m,n) = (0,0)$. Here, `.' represents the dot product operation, and $\lambda_{1,2}$ are hyper-parameters.
    
\end{enumerate}
********************************************************************************************************
\section{Experiments}
In this section, the implementation details, ISR comparison results, and the effect of the regularization term of the proposed NSSR-DIL method were discussed.
\subsection{Training setup}
We trained the proposed L-CNN (refer Sec.\ref{L-CNN} for architecture details of L-CNN) on the RKG dataset (refer Sec. \ref{kdataset}) with the learning objective given in Eq. (\ref{eq6}). The number of epochs was $50$ and the learning rate was $0.1$ with a step scheduler. The Adam \cite{adam} optimization was used, with $\beta =0.9$. The values of hyper-parameters used in the learning objective i.e., Eq. (\ref{eq8}), set empirically, are as follows, $\lambda_1 = 0.8,~\lambda_2 = 0.2$. We have used a computer with an NVIDIA-GTX 2080 Ti 11GB GPU in all our experiments.
\subsection{Results}
\label{sec:results}
\textbf{Super-Resolution.} The ISR ability of our approach was evaluated on the benchmark datasets RealSR \cite{cai2019toward} for sf 2, sf 3, sf 4 and DIV2KRK \cite{kernelgan} for sf 2, sf 4. The results are outlined in Table \ref{tab:srresults}. For a fair comparison, we considered works like KernelGAN \cite{kernelgan}, ZSSR \cite{zssr}, DBPI \cite{dbpi}, and DualSR \cite{dualsr} which consider single input LR image only for modeling the ISR task as in our experiments. Besides, we included the recent SotA supervised ISR methods Dual Aggregation Transformer (DAT) \cite{dat}, and MIRNetV2 \cite{mirnetv2}, in our comparison experiments for adequate validation. We note that the proposed NSSR-DIL is only the DL approach in the literature that learns the inverse degradation kernel from the degradation kernel itself, without the need for LR image input. While, the compared SotA methods need supervised learning from images for $K$ estimation and/or ISR model learning.
\\ \textbf{Computational complexity}. The proposed L-CNN is an efficient, lightweight ISR model. The comparison of computational complexity in terms of parameters and super-resolution time (inference time) in minutes with standard self-supervised and zero-shot methods is given in Table \ref{tab:srresults}. The proposed NSSR-DIL requires 15 times fewer memory resources even when compared with the least among the SotA methods i.e. ZSSR. Besides, the inference time required for the NSSR-DIL is reduced to less than a minute and less by an order of 100s than compeer works. Here, the recent SotA methods are excluded from the complexity comparison experiments that employ sophisticated deep models trained on large ISR datasets.
\par
\begin{figure}[]
    \centering
    \includegraphics[width = 0.7\textwidth, height = 4cm]{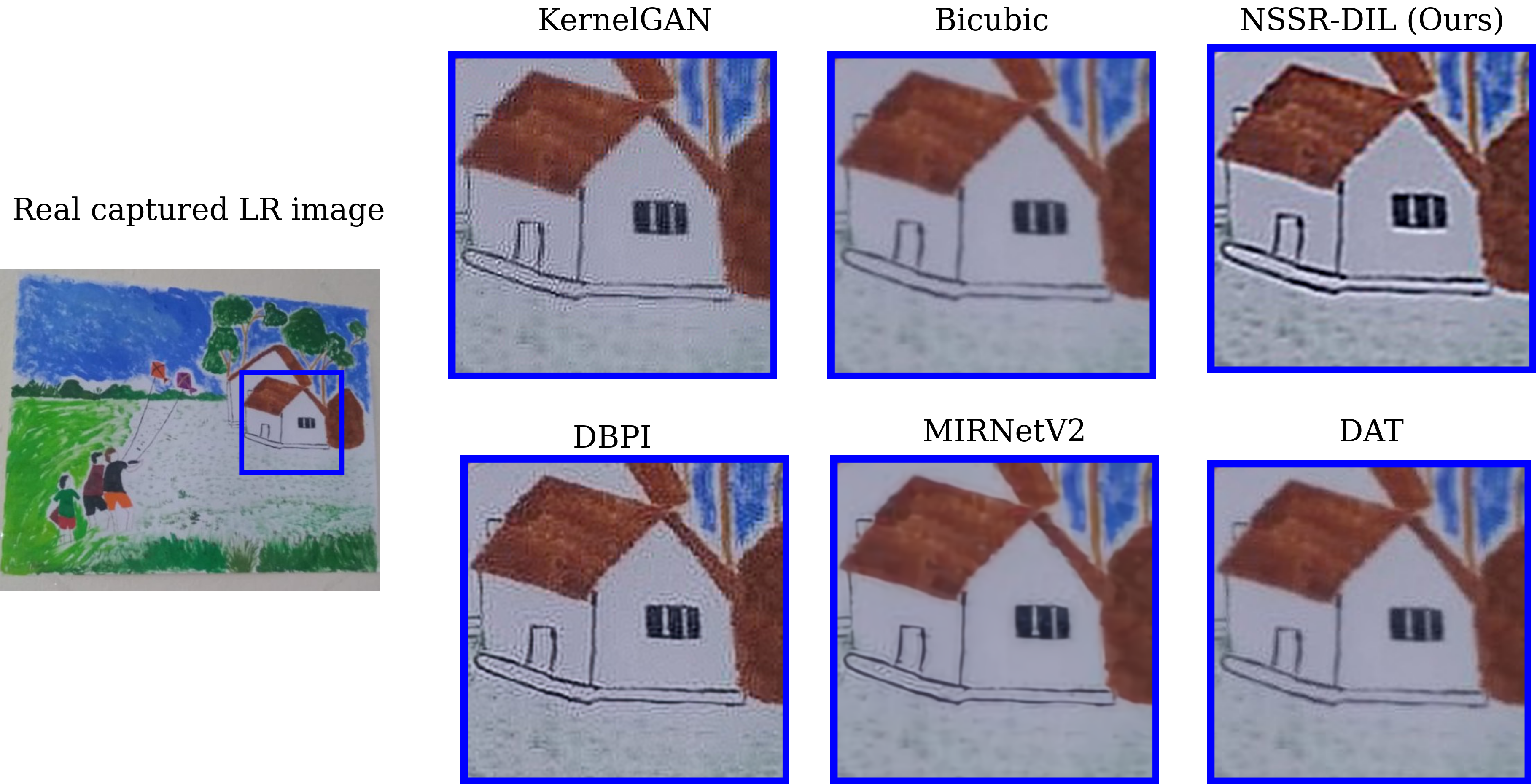}
    \caption{Visual results of the ISR of the cropped regions of a real captured image using Samsung J7 Prime (13MP, f/1.9) for the SR sf 2.}
    \label{fig:realresult}
\end{figure}

\begin{figure*}[!h]
    \centering
    \includegraphics[width = 0.95\textwidth, height = 8.5cm]{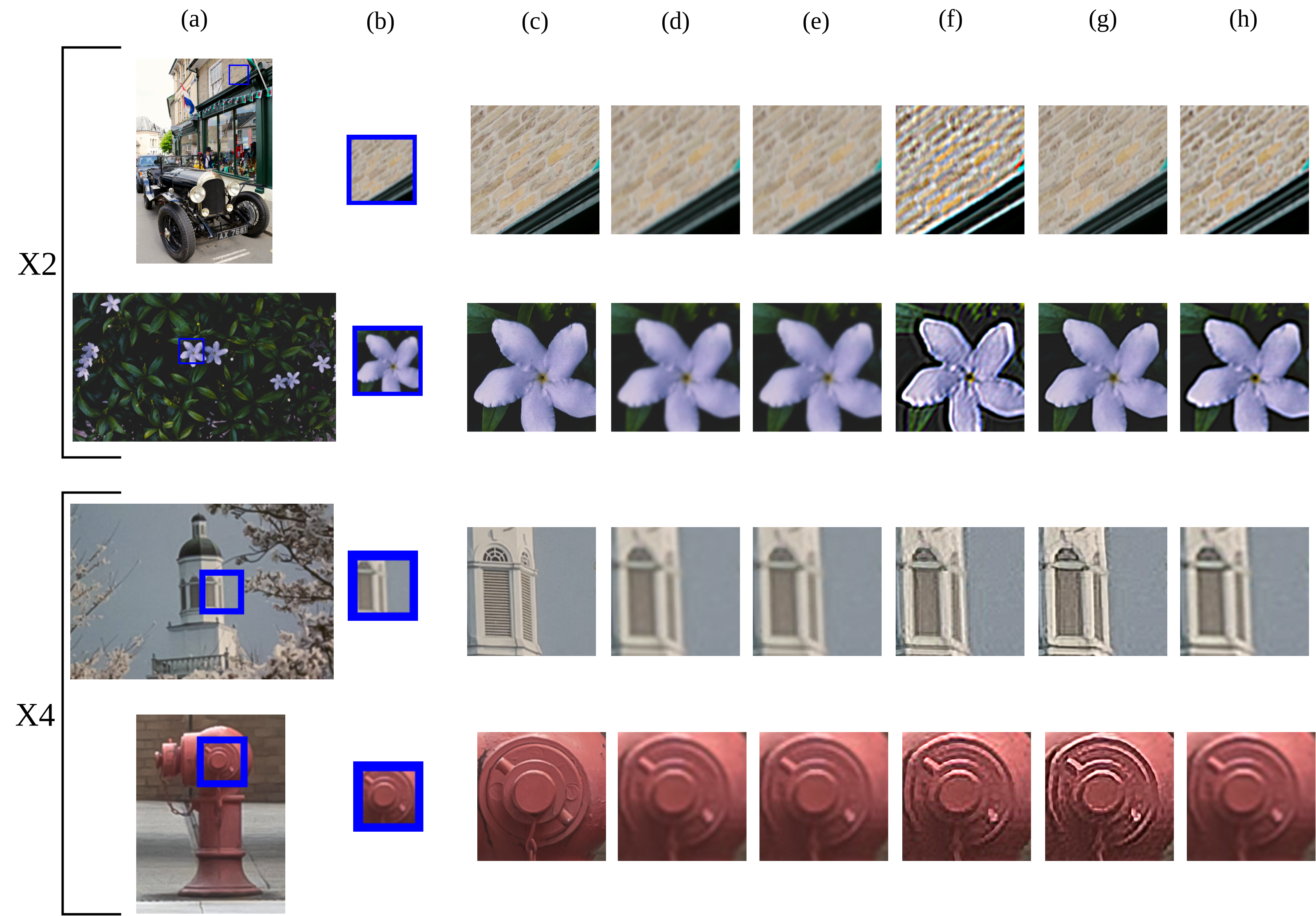}
    \caption{Qualitative results of different ISR methods for the scale factor 2 \& 4 using DIV2KRK and RealSR datasets respectively. Here (a) - LR image, (b) - LR patch, (c) - HR image, (d) - Bicubic, (e) - ZSSR, (f) - KernelGAN + ZSSR,  (g) - DBPI, (h) - NSSR-DIL.}
    \label{fig:Visual-Div-x2}
\end{figure*}

 In practice, for real test instances, the reference image is not available and the primary interest is to generate the HR version of the given LR image close to the natural image statistics. Hence, we considered Neural Image Assessment (NIMA) \cite{NIMA}, a no-reference image quality assessment metric with a high correlation to human perception, to demonstrate the performance of our proposed method. Besides, to assess the ability of the proposed NSSR-DIL to restore the finer details during SR processes, we employ Structural Similarity Index Measure (SSIM), and for qualitative and quantitative similarity with the HR image we provide the visual results and Peak Signal to Noise Ratio (PSNR) values. Despite being very lightweight the proposed NSSR-DIL demonstrated its effectiveness significantly in two out of three standard metrics on real and synthetic ISR datasets. Since the proposed model did not train on either LR or LR-HR image pair data, a relative decline in reference-based metric at pixel level i.e., PSNR was observed as expected. The important note is that the proposed model did not produce artifacts in the generated super-resolved output images, unlike the compeer, SotA works like \cite{kernelgan}, \cite{dbpi}, \cite{mirnetv2}. The sample visual results were provided in  Fig. \ref{fig:Visual-Div-x2}.\\ 
We would like to emphasize that the RealSR dataset contains real LR images with unknown noise levels and degradation kernel information \cite{realsr}. Furthermore, the DIV2KRK dataset includes synthetically generated LR images degraded by the Gaussian kernels perturbed by uniform multiplicative noise \cite{kernelgan}. Even though we didn't consider the additive noise explicitly in our degradation model represented by Eq. \ref{eq1}, the proposed model showcased its ability to generate super-resolved images with superior SSIM and NIMA scores for both datasets. This highlights the ability of the proposed NSSR-DIL model to be robust to unseen degradations and to generate natural images with fine details in ISR.\\
\textbf{{ISR of real captured LR image}}. Additionally, the comparison of ISR performance on RLR image provided in Fig. \ref{fig:realresult}, showcases that our method super-resolves with essential fine details, and without introducing artifacts, unlike its compeers. Here, the noise distribution and the degradation kernel information are unknown. Thus our ISR method demonstrated its generalizability without any knowledge of the corresponding blur kernel, has no strict limitations on the blur kernel estimation, and is also not restrictive to the assumptions considered in our degradation kernel estimation. 
\begin{table*}[!t]
\large
\centering
\caption{Comparison of the computational complexity of SotA ISR methods and their performance on RealSR and DIV2KRK datasets in terms of NIMA$\uparrow$/SSIM$\uparrow$/PSNR$\uparrow$ were given below. Here, the \textcolor{red}{red} indicates the best score and the \textcolor{blue}{blue} indicates the second-best score.}
\label{tab:srresults}
\resizebox{0.975\textwidth}{!}{%
\begin{tabular}{ccccccc|cc}
\hline 
 &
  &
  
  \textbf{ZSSR} &
  \textbf{\begin{tabular}[c]{@{}c@{}}KernelGAN\\ + ZSSR\end{tabular}} &
  \textbf{DBPI}&
  \textbf{DualSR} & 
  \textbf{\begin{tabular}[c]{@{}c@{}}NSSR-DIL\\   (Ours)\end{tabular}}  &  
   \textbf{MIRNetV2} &
  \textbf{DAT}
   \\\hline
   \textbf{\begin{tabular}[c]{@{}c@{}}No.of \\ Parameters\\ (Million)\end{tabular}} &
   &
   
  \textcolor{blue}{0.29} &
  0.151+0.29 &
  0.5 &
  0.45&
  \textcolor{red}{0.028}
  &
    N.A. &
 N.A. 
   \\
  
  \hline 
\textbf{\begin{tabular}[c]{@{}c@{}}Inference \\ time (min)\end{tabular}} &
&

  $>=$10 &
  $>=$13 &   
 \textcolor{blue}{ $>=$1}  &
  $3.5$ &
 \textcolor{red}{0.005} 
 &
  N.A. &
 N.A.  \\
  \hline \hline \\

\multirow{3}{*}{\textbf{RealSR}} &
  \textbf{x2} &

  3.996/0.8786/\textcolor{red}{30.56} &
  3.81/\textcolor{red}{0.8907}/\textcolor{blue}{30.24} &
  3.909/0.8226/27.83  &
  \textcolor{red}{5.0020}/0.8570/28.01 &
  \textcolor{blue}{4.075}/\textcolor{blue}{0.8836}/25.29 &
   {4.9329}/0.8625/29.94 &
 {4.9413}/0.8709/{30.30}
   \\ \\ \cline{2-9}  \\
 &
  \textbf{x3} &
  
  \textcolor{blue}{4.008}/\textcolor{blue}{0.6210}/\textcolor{blue}{20.53} &
  -/-/- &
  -/-/- &
  -/-/- &
  \textcolor{red}{4.095}/\textcolor{red}{0.8395}/\textcolor{red}{24.85} &
   {4.9372}/0.7765/26.97 &
{ 4.9322}/0.7884/{27.29}  \\ \\ \cline{2-9} \\
 &
  \textbf{x4} &

  4.058/\textcolor{blue}{0.7434}/\textcolor{red}{25.83} &
  \textcolor{blue}{4.068}/0.7243/24.09 &
  4.020/0.6508/22.21 &
  -/-/- &
  \textcolor{red}{4.118}/\textcolor{red}{0.8079}/\textcolor{blue}{24.15}&
 { 4.9273}/0.7283/{25.61} &
  {4.8973}/0.1211/11.69\\ \\ \hline \\
\multirow{2}{*}{\textbf{DIV2KRK}} &
  \textbf{x2} &

  4.111/0.7925/27.51 &
  4.071/0.8379/{28.24} &
  4.049/\textcolor{red}{0.8684}/\textcolor{red}{30.77} &
 \textcolor{red}{5.005}/{0.8538}/\textcolor{blue}{29.38} &
  \textcolor{blue}{4.161}/\textcolor{blue}{0.8644}/26.02 &
{5.020}/0.7845/{26.85} &
 {4.9676}/{0.8046}/{27.81}  \\ \\ \cline{2-9} \\
 &
  \textbf{x4} &
  
  4.156/0.6550/24.05 &
  4.089/0.6799/\textcolor{blue}{24.76} &
  4.146/\textcolor{blue}{0.7368}/\textcolor{red}{26.86}& -/-/- &
 \textcolor{blue}{ 4.170}/\textcolor{red}{0.7926}/23.58  &
  {5.001}/0.6991/{24.92} &
{4.9081}/0.6631/24.23  
   \\ \\ \hline 
\end{tabular}%
}
\end{table*}

\subsection {Ablation study}
In this section, the significance of the proposed regularization term ($R$) (given in Eq. \ref{eq8}), the performance of the NSSR-DIL method for various sizes of the RKG dataset, and the performance of the NSSR-DIL model on unseen synthetic test dataset were discussed.
\subsubsection{Regularization term (R)}
In the proposed learning objective (refer Eq. \ref{eq6}), we introduced two constraints through the regularization term ($R$) (refer Eq. \ref{eq8}) to obtain the inverse degradation model and with the reliable ISR performance. The influence of each entity in the proposed $R$ is quantified and presented in Table \ref{tab:R}. It was observed that the presence of $L_{ConvArea}$ in $R$ has shown a greater impact on the performance of the proposed method relative to the $L_{center}$. The term $L_{center}$ influenced the perceptual quality of the images. The term $L_{ConvArea}$ together with the $L_{center}$ assisted the NSSR-DIL model to attain a reliable performance.\\
\textbf{Note:}  The ISR performance results of the NSSR-DIL method for various sizes of the RKG dataset, and also on unseen synthetic test dataset are provided in the supplementary material.
\begin{table}[!h]
\centering
\Large
\caption{The effect of each of the proposed regularization terms in R on the NSSR-DIL model's performance with the DIV2KRK dataset for sf 2.}
\label{tab:R}
\resizebox{\textwidth}{!}{%
\begin{tabular}{|l|l|l|l|l|}
\hline 
\textbf{Loss (L) } &       $ \lVert K \ast K^{-1} - ~\delta \rVert^2_2 $             &     \begin{tabular}[c]{@{}c@{}}$\lVert K \ast K^{-1} - ~\delta \rVert^2_2 + $\\$  \lambda_1 \times L_{ConvArea}$  \end{tabular}                &    \begin{tabular}[c]{@{}c@{}}$\lVert K \ast K^{-1} - ~\delta \rVert^2_2 +  $\\$ \lambda_2 \times L_{Center}$      \end{tabular}          &      \begin{tabular}[c]{@{}c@{}} $ \lVert K \ast K^{-1} - ~\delta \rVert^2_2 + $\\ $ \lambda_1 \times L_{ConvArea} + \lambda_2 \times L_{Center}$      \end{tabular}    \\      \hline 
\textbf{NIMA$\uparrow$}/\textbf{SSIM$\uparrow$}/\textbf{PSNR$\uparrow$} & 3.9138/0.0153/6.58 & 4.1550/0.8096/24.22 & 3.099/0.6360/12.69 & 4.161/0.8644/26.02 \\ \hline
\end{tabular}
}
\end{table}
\section{Conclusion}
We have established a new problem formulation in terms of DIL for the ISR task. The proposed computationally efficient NSSR-DIL is the first image data-independent DL-based ISR model, for any scale factor. The proposed method's performance is quite comparable to the SotA works, despite being independent of image data in the ISR model design. Thus, this work demonstrates tremendous hope for improving the ISR capability without the need for image datasets (i.e., both supervised and unsupervised ISR datasets). Our NSSR-DIL method paves the path to have a deeper look into the learning and understanding of the inverse degradation model from a linear systems perspective. The experimental results and computational performance comparisons with the SotA indicate that the proposed NSSR-DIL are remarkably suitable for real-time ISR tasks and embedded ISR applications.
\bibliography{egbib}
\end{document}


\maketitle
\section {Ablation study}
In this section, the performance of the NSSR-DIL method for various sizes of the RKG dataset, and on unseen synthetic test dataset were quantified and discussed. 
\subsection{Image Super-Resolution of scale factors $ > \times 4$}
The existing SotA ISR works are limited to performing the ISR task for scale factors up to 4, especially the zero-shot approaches. The present zero-shot approaches are restricted by their assumption in the ISR task definition i.e. patch recurrence property across the scales of the image, to extend for sfs beyond $\times 4$ \cite{zssr,dbpi} etc. Additionally, the existing SotA ISR methods based on deep models are bound to increase the computational complexity in terms of memory and time to achieve sfs above $\times 4$. 
\par In this work, the proposed NSSR-DIL model can be easily extended to any higher scale factors like  $\times 16 ~\& \times 32$ without an increase in computational footprint. The performance of the proposed NSSR-DIL model was found satisfactory when compared to traditional interpolation methods like Bicubic interpolation, which can also be extended to any sfs. It is crucial to note that, in the case of NSSR-DIL, the number of parameters remains the same irrespective of varying sf.\\ 
In our experiments, the HR ground-truth images in the DIV2KRK dataset were considered to generate the LR images of $\times 8, \times 16, \times 32$. The anisotropic gaussian kernels having $\sigma_1, \sigma_2 \in U[0.175, 3.1]$ and rotation angle $\theta \in U[0, \pi]$ with dimensions $31 \times 31, 41 \times 41, 51 \times 51$ were used to generate the LR images of sf $\times 8, \times 16, \times 32$ respectively. The results were tabulated in Table no. \ref{tab:highersfs}. It was observed that the proposed method superseded the standard Bicubic interpolation method in terms of SSIM for higher sfs.

\begin{table}[]
\centering
\caption{Comparison of ISR performance of higher scale factors using SSIM$\uparrow$/PSNR$\uparrow$ metrics.}
\label{tab:highersfs}
\resizebox{0.55\textwidth}{!}{%
\begin{tabular}{ccc}
\hline 
\textbf{Scale factor} &
  \textbf{Bicubic} &
  \textbf{\begin{tabular}[c]{@{}c@{}}NSSR-DIL\\ (Ours)\end{tabular}}  \\ \hline
x2  & 0.7846/27.24 & 0.8644/26.02  \\ \hline
x4  & 0.6478/23.89 & 0.7926/23.58 \\ \hline
x8  & 0.6250/23.31 & 0.7625/23.35 \\ \hline
x16 & 0.5666/21.33 & 0.7132/21.34 \\ \hline
x32 & 0.5348/19.34 & 0.6713/19.25 \\ \hline
\end{tabular}%
}
\end{table}


\subsection{Performance of NSSR-DIL on unseen synthetic dataset}

We have prepared a synthetic test dataset with the degradation kernels outside of our training dataset. We used the DIV2K validation set to generate LR images using random Gaussian kernels with size $11 \times 11$ and $21 \times 21$ of varying shapes, and orientations i.e. $\sigma_1, \sigma_2 \in U[3, 5]$ and rotation angle $\theta \in U[0, \pi]$  for scale factor 2 and 4 respectively. These random Gaussian kernels are further perturbed by uniform multiplicative noise as well. The model’s performance results on this synthetic ISR dataset in Table \ref{tab:unseen} demonstrate the robustness of the proposed NSSR on unseen degradations. 

\begin{table}[!h]
\centering
\caption{ISR performance of the proposed NSSR-DIL on the synthetic dataset generated using the unseen kernels in terms of SSIM$\uparrow$, NIMA$\uparrow$, and PSNR$\uparrow$ were given below}
\label{tab:unseen}
\resizebox{0.65\textwidth}{!}{%
\begin{tabular}{|l|l|l|l|l|}
\hline
 & \textbf{Scale factor} & \textbf{SSIM} & \textbf{NIMA} & \textbf{PSNR} \\ \hline
\multirow{2}{*}{\textbf{NSSR-DIL}} & {X2} & {0.8257} & {4.94} & {24.79} \\ \cline{2-5} 
 & {X4} & {0.7741} & {4.88} & {23.25}\\ \hline
\end{tabular}%
}
\end{table}

\subsection{Performance of NSSR-DIL with varied sizes of RKG dataset}
\begin{figure*}[!h]
    \centering
    \includegraphics[width = 0.825\textwidth, height = 3cm]{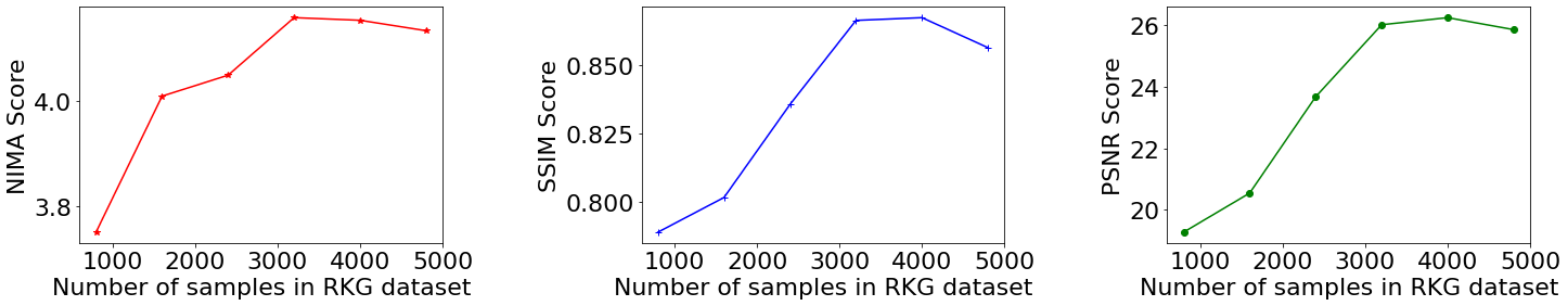 }
    \caption{The performance comparison of the NSSR-DIL model with sizes of RKG dataset.}
    \label{fig:samples}
\end{figure*}

We studied the performance of the NSSR-DIL model trained on varied sizes of the RKG dataset. In our study, we generated five different datasets with the number of samples $800, 1600, 2400, 3200, 4000$, and $4800$ and named as $RKG_n$, where $n = 1, 2, 3, 4, 5$, respectively. These five datasets i.e. $RKG_n$ were generated following the steps discussed in Sec. 3.2. The L-CNN was trained independently on each set and evaluated on the DIV2KRK test dataset for sf 2. The visual plots depicting the L-CNN performance comparison for $RKG_n$ vs evaluation metrics i.e. NIMA, SSIM, and PSNR are presented in Fig. \ref{fig:samples}. It is observed that there is a linear improvement in these metrics as the number of samples was increased initially and saturated later. Therefore we considered 3200 training samples i.e., $RKG_4$ dataset in all our experiments.

\bibliography{egbib}